\documentclass{svproc}
\usepackage{url}
\usepackage[ruled,vlined]{algorithm2e}
\usepackage{amsmath}
\usepackage{amssymb}
\usepackage[FIGBOTCAP]{subfigure}
\usepackage{multirow}
\usepackage{graphicx} 
\usepackage{epsfig}
\usepackage{comment}

\begin{document}
\mainmatter

\title{Integrated Modular Solution for Task Oriented Manipulator Configuration Design}

\titlerunning{Task Oriented Modular Composition Design}
\author{Anubhav Dogra \and Sakshay Mahna \and Srikant Sekhar Padhee \and Ekta Singla}

\authorrunning{Anubhav Dogra et. al.}

\institute{Indian Institute of Technology Ropar, Rupnagar, Punjab, India 140001\\
\email{2016mez0019@iitrpr.ac.in, 2018csb1119@iitrpr.ac.in, sspadhee@iitrpr.ac.in, ekta@iitrpr.ac.in}}

\maketitle
\begin{abstract}
	Modular and reconfigurable robotic systems have been designed to provide a customized solution for the non-repetitive tasks to be performed in a constrained environment. Customized solutions are normally extracted from task-based optimization of the possible manipulator configurations but the solution are not integrated, for providing the modular compositions directly. In this work, in the first phase, a strategy of finding unconventional optimal configurations with minimal number of degrees-of-freedom are discussed based upon the prescribed working locations and the cluttered environment. Then, in the second phase, design of the modular and reconfigurable architecture is presented which can adapt these unconventional robotic parameters. Rather than generating and evolving the modular compositions, a strategy is presented through which the unconventional optimal configurations can be mapped directly to the modular compositions. The generated modular composition is validated using Robot Operating System for the motion planning between the prescribed working locations in a given cluttered environment.      
\end{abstract}

\section{Introduction}
Providing the robotic solutions for the custom tasks needs time and cost-effective customizable systems. These robotic systems should be adaptable to all the variations of the prescribed tasks and the constrained working locations where the standard robotic configurations cannot work. These tasks may belong to industry such as pick and place works in highly cluttered environments, in servicing or maintenance sectors, or in health care sectors~\cite{singh2018modular}. Given an environment and the working locations, a robotic solution strategy has been provided by the various researchers - using modularity and reconfiguability, but is never integrated. Attention is required to provide a modular solution which can directly give the modular configurations based upon the prescribed task~\cite{dograJMD2021}. Configuration synthesis has been explored by many reseachers in recent years providing different frameworks considering various factors such as reachability, kinematic and dynamics performances, obstacle avoidance, path planning etc. Campos et. al~\cite{campos2019task} presented configuration synthesis for reachability in 3D space with obstacle avoidance using Sketch synthesis system and used HEBI actuators for their modular configurations. Icer et. al.~\cite{icer2017evolutionary} uses evolutionary algorithms to generate possible modular compositions and to find their best modular composition for given task and environment. Similar work is presented by Liu et. al.~\cite{liu2020optimizing} for optimizing modular composition in static environment, also considering energy efficiency, cycle time and few performance metrics.  Kereluk et. al.~\cite{kereluk2017task} developed a reconfigurable 12-Degree-of-Freedom (DoF) manipulator and optimizes the number of joints to be active or locked for the prescribed task. Stravopodis et. al.~\cite{stravopodis2020rectilinear} presented 3-DoF modular design with psuedo joint and optimizes the configuration for a recitlinear task using Genetic Algorithm. Whitman et. al.~\cite{whitman2018task} presented a motion planner which also synthesizes the manipulator for required tasks which includes elimination of the joints to reduce the DoF if required. Along with that, various modular designs has been presented in literature to provide the customized solutions and can be seen in~\cite{chen2016modular,mohamed2017pose,Modman2020,valente2016reconfigurable,singh2016realization,dograJMD2021} \\
These solutions are majorly explored on the basis of generating modular compositions from a given set of modules and then checking the configurations if able to work or not on the prescribed tasks. Alongside, the configurations are explored mostly for a fixed DoF system and standard parameters. In this paper, we have presented: 1. An optimal and unconventional configuration synthesis of the manipulator along with minimizing the number of DoF required for the prescribed working locations and a cluttered environment. 2. Design of a modular library which can adapt to the unconventional parameters in the optimal configurations. 3. Integration of optimal synthesis with modular compositions, which provides the optimal modular sequence having two variants of the modules as Heavy (H) and Light (L).\\
The paper is organized as follows. Section~\ref{sec:config_synthesis} defines the approach and methodology for the optimal configuration synthesis in a given environment and specified working locations. Section~\ref{sec:module design} presents the design of each unit of the module and its features. Section~\ref{sec:modularconfigurations} shows the rules of assembling different configurations considering the designed modules, and composition possibilities. Section~\ref{sec:results} discusses about the outcomes of the approach, with the conclusion of the work in last section.
\section{Optimal Configuration Synthesis: Minimized DoF Approach}\label{sec:config_synthesis}
The objective of minimizing number of DoF is considered in this paper to determine the best possible configuration of a manipulator to do a specific set of tasks. 
A nested bi-level optimization problem has been formulated, in which first level is used to compute the DoF of the manipulator and the second level provides the optimal robotic parameters. Binary search method is applied for the computation of DoF as a unidirectional search problem during optimization at the upper level~\cite{singh2018modular}.
Algorithm~\ref{algo:1} shows the steps of computation, where $k_l$ and $k_u$ are the lower and the upper limits of the element number in the given array, say $A$. 
\begin{algorithm}[]
	\SetAlgoLined
	\KwResult{n=DoF}
	initialization: $k_l$= $0$; $k_u$= $k-1$\;
	\While{$k_l$ $\leq$ $k_u$}{
		$m$ = floor(($k_l$ + $k_u$) / $2$); n=$A[m]$\;		
		\eIf{$f$ $\leq$ T}{
			$k_u$= $m-1$\;
		}{
			$k_l$= $m+1$\;}
	}
	\caption{Binary search algorithm to initialize DoF for configuration synthesis}
	\label{algo:1}
\end{algorithm}
$A$ contains $k$ number of elements as  $\{A_0,~A_1,~A_2,~.~.~.,~A_{k-1}\}$. $f$ is the optimization objective function value and $n$ is the number of DoF. The result from the outer problem of binary search goes into the (inner) lower level and iterations occur accordingly with respect to the function value.  
\subsection{Constrained Optimization Formulation}\label{sec:opti}
The desired manipulator is required to reach the set of Task Space Locations (TSLs) in a cluttered environment. Thus optimization problem is formulated to design an optimal configuration with minimum number of DoF which can reach to the TSLs avoiding obstacles. $P_d = [x_d,~y_d,~z_d]$ is considered as desired position vector, and $O_d = [\alpha_d,~\beta_d,~\gamma_d]$ as the desired orientation vector of the end-effector frame or last frame. Thus, for $N$ number of TSLs, the objective function can be derived as the error between the desired and the current pose.
\begin{eqnarray}
P_{error}& = &\sum_{i=1}^{N} [(x_{id}-x_{ia})^2 + (y_{id}-y_{ia})^2 + (z_{id}-z_{ia})^2];\\
O_{error}& = &\sum_{i=1}^{N} [(\alpha_{id}-\alpha_{ia})^2 + (\beta_{id}-\beta_{ia})^2 + (\gamma_{id}-\gamma_{ia})^2];\\
f(\textbf{x})& = &|\sqrt{P_{error}}|+|\sqrt{O_{error}}|;
\end{eqnarray}  
Here, subscript $a$ denotes actual parameters of position and orientations computed through homogenous transformations at the current design variables as
\begin{equation}
T_n^0 =  T_1^0 ~T_2^1.~.~.~T_n^{n-1}.
\end{equation}  
The configuration of a manipulator is defined by a set of Denavit-Hartenberg (DH) parameters, which are all considered as variables in this study. For $N$ number of TSLs in a Cartesian space, the manipulator with same robotic parameters is required to reach all TSLs. Thus, the total number of design variables would be $(3+N)*n$ as $[a_i,~\alpha_i,~d_i,~\theta_{ij}]$, where $i~\in~1:n$, and $j~\in~1:N$.

To avoid collisions with obstacles, the obstacles and the links are assumed as triangulated meshes. Each link is assumed to be a parallelepiped with a certain width. The minimum distance $(D)$ between the two meshes is computed and treated as a constraint. Collision check is done among links and between the links and the obstacles. Value of the $D$ becomes negative when there is collision between the meshes. Thus non-linear inequality constraint can be written as
\begin{equation}
g_{ij}(\textbf{x})=-D_{ij}-\delta \leq 0;~ i \in 1:n, ~j \in 1:N
\end{equation}  
where $\delta$ can be used as a safety margin to avoid just touching of links with the obstacles.
Thus the optimization model can be written as.\\

To Minimize:
\begin{eqnarray}\label{eq:objective_function}
f(\textbf{x})& = &|\sqrt{P_{error}}|+|\sqrt{O_{error}}|;
\end{eqnarray}

with design variables vector,
\begin{equation}\label{eq:DV}
\textbf{x} = [a_{i},\alpha_{i},d_{i},\theta_{ij}]^T,for~i \in 1:n, j\in 1:N.
\end{equation}

Subjected to:
\begin{eqnarray}\label{eq:constraints}
a_i^l \leq &a_i \leq &a_i^u,\\
\alpha_i^l \leq &\alpha_i \leq &\alpha_i^u,\\
d_i^l \leq &d_i \leq &d_i^u,\\
\theta_{ij}^l \leq &\theta_{ij} \leq &\theta_{ij}^u,\\
-D_{ij}-&\delta \leq& 0
\end{eqnarray}

\section{Modular Architecture}\label{sec:module design}
\begin{figure} 
	\centering
	\subfigure[]{\includegraphics[width=1in]{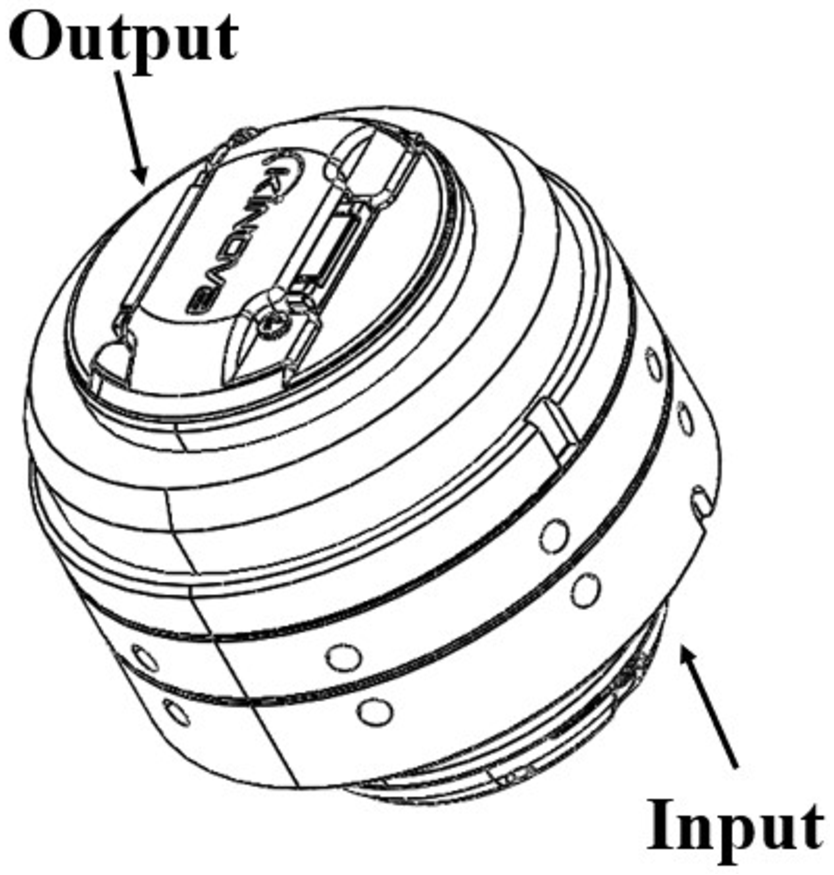}
	}
	~
	\subfigure[]{\includegraphics[width=2in]{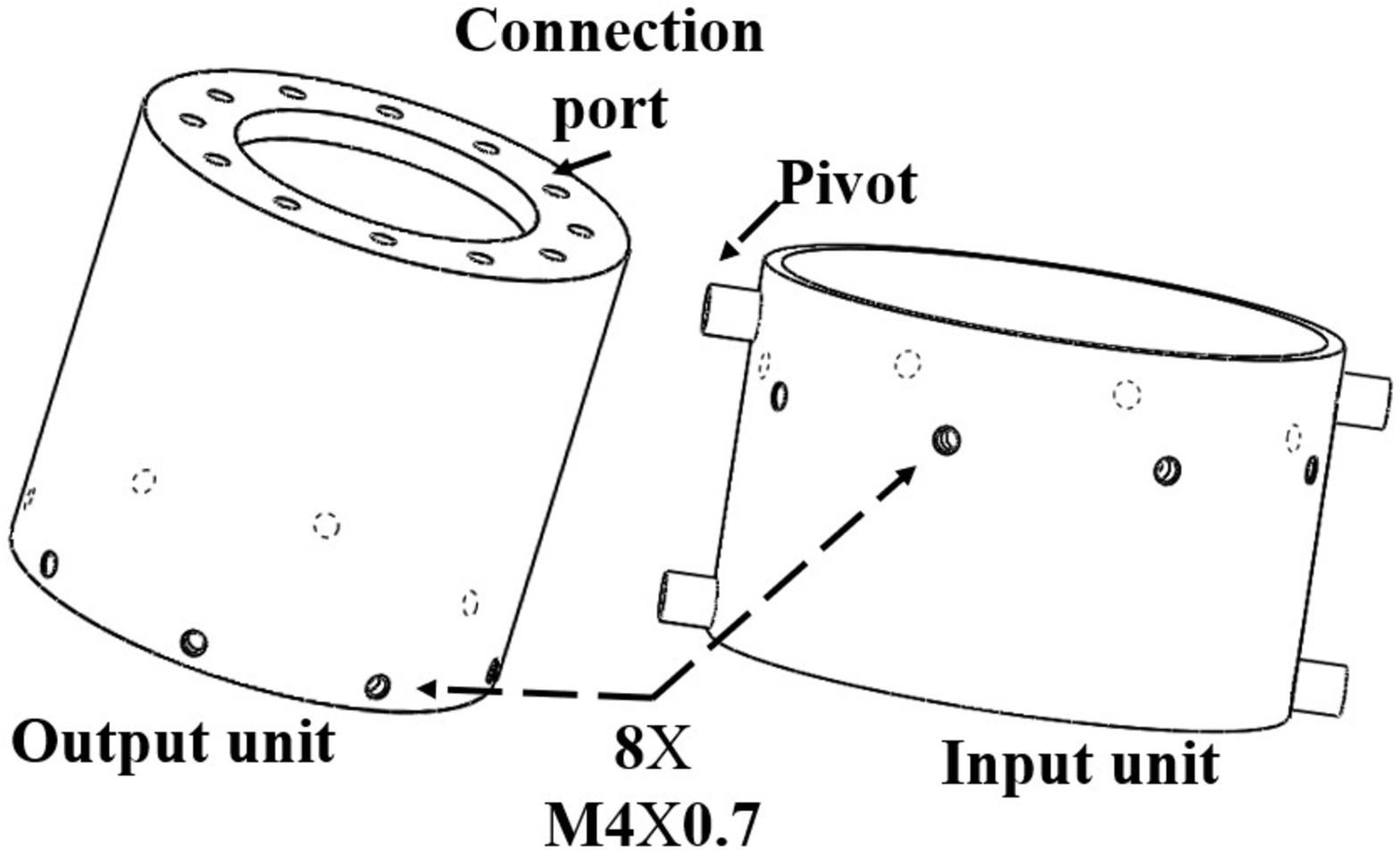}
	}
	~
	\subfigure[]{\includegraphics[width=1.75in]{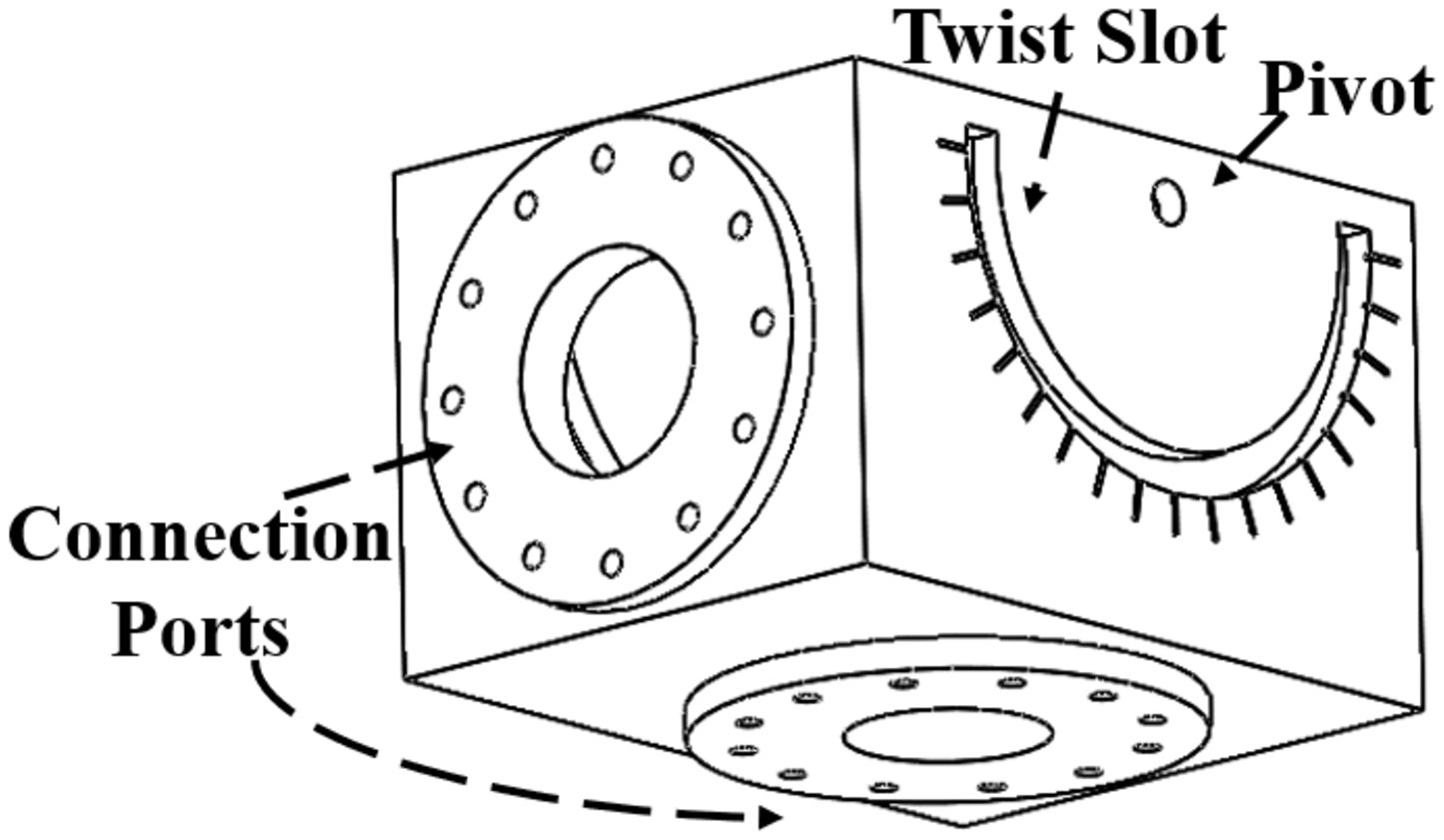}
	}
	~
	\subfigure[]{\includegraphics[width=1in]{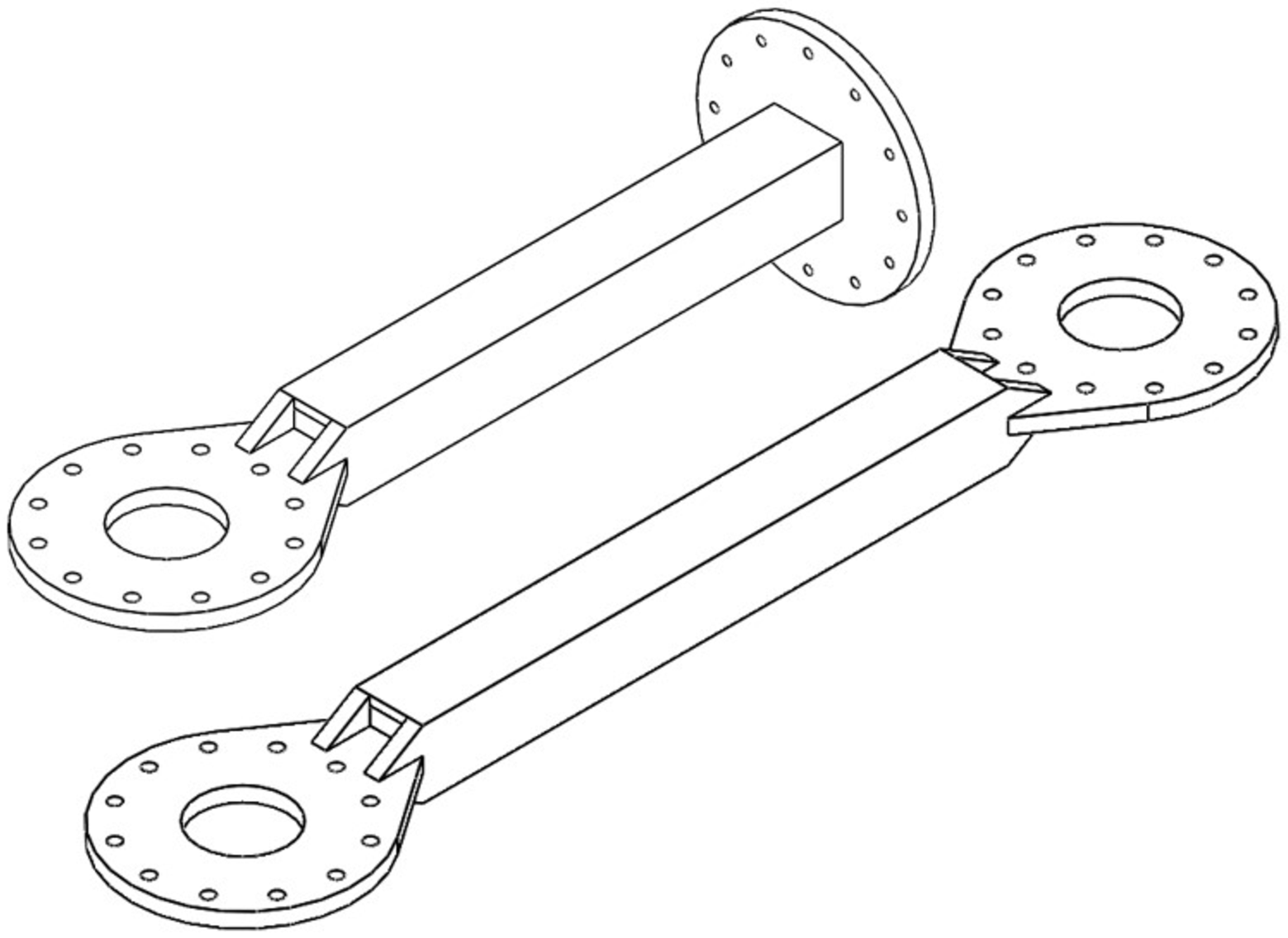}
	}
	\caption{Modular library containing (a)  actuator, (b) actuator casings (c) Adaptive twist unit, (d) Link modules.}
	\label{fig:joint_unit}
\end{figure}

The presented modular library~\cite{dograJMD2021} mainly consists of link modules and joint modules. The joint module is designed to incorporate the actuator and its adjustment in terms of connection with other modules and the required arrangement of twist angles. It consists of mainly $3$ components as shown in Fig.~\ref{fig:joint_unit}, an actuator, an unconventional twist unit, and two casings for input and output section of the actuator. To design the modular joint assembly, \textit{KA-Series} actuators from \textit{Kinova}~\cite{Kinova2019} are used in this work.
Specifications of both the actuators are provided in Table~\ref{tab:actuator_specs}. $\epsilon$ is the parameter which defines the capacity of the designed modules to pick up the the number of modules of its kind. This study was done by the authors based on worst-torque analysis of the modular assemblies in~\cite{dograJMD2021}. Link modules are passive modules to provide structural aspects to a manipulator. Two kinds of links are designed as shown in Fig.~\ref{fig:joint_unit} (d). One type has the parallel connection ports and the other has perpendicular.. 
\begin{table}[t]
	\caption{Technical specifications of actuators}
	\begin{center}
		\label{tab:actuator_specs}
		\begin{tabular}{l c c}
			& & \\
			\hline
			\hline
			Quantity & KA-75+ & KA-58 \\
			\hline
			\hline
			Weight (kg) & 0.57 & 0.357 \\
			No load speed (rpm)& 12.2 & 20.3 \\
			Nominal torque (Nm)& 12 &  3.6 \\
			Maximum torque (Nm)& 30.5 &  6.8 \\		
			$\epsilon$ & 3 & 3\\	
			\hline
			\hline
		\end{tabular}
	\end{center}
\end{table}

Adaptive twist unit is designed to incorporate the twist angles adjustments between two adjacent frames according to \textit{DH} convention. The twist adjustment can be done in two ways. One is the angle between the two adjacent intersecting joint axes, another is an angle between two adjacent skew joint axis. In this design, pivotal extension from the input unit of casing is assembled with pivot of twist adjusting unit, see Fig.~\ref{fig:joint_unit}. The twist unit has a semi-circular slot with resolution of $10^\circ$ ranges from $0$ to $+90^\circ$ and $-90^\circ$ in both clockwise and anticlockwise direction as shown in Fig.~\ref{fig:joint_unit} (c). The input unit can rotate about pivotal axis in the twist unit to adjust the twist angle. Besides, there are two connection ports provided orthogonal to each other and are provided with $12$ holes with resolution of $30^\circ$ to adjust twist angle. \\
Two casings are designed for an actuator for each of its sides. The input casing is to be assembled with input section of actuator, which is relatively fixed. Thus, it has two pivotal extensions, as shown in Fig.~\ref{fig:joint_unit} (b), which are assembled with the twist unit. The output casing is to be assembled with the output section which rotates with respect to the input section. On each of the casing, there are eight tapped holes which are used to fix these casings onto the actuator. Output casing has $12$ threaded holes. These $12$ holes gives the resolution of $30^\circ$ when assembling other modules to it.

Post analysis of the design of the modules including validations, modular assembly combinations and possibilities, worst torque analysis are not given in this paper, and can be checked in authors work in~\cite{dograRobotica2021,dograJMD2021}. 
\section{Modular Composition based upon Optimal Configuration}\label{sec:modularconfigurations}
\begin{figure} 
	\centering
	\subfigure[]{\includegraphics[width=0.75in]{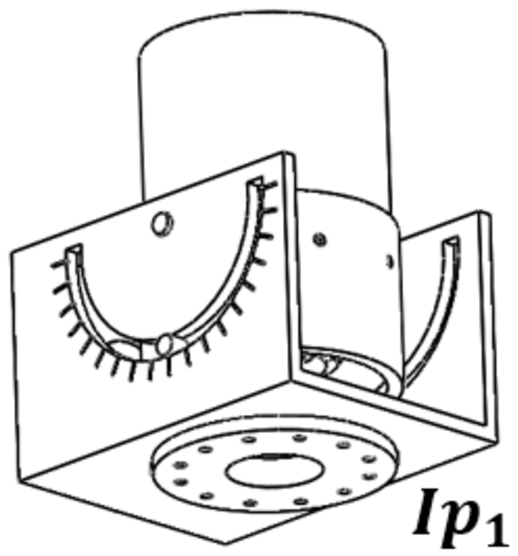}
	}
	~
	\subfigure[]{\includegraphics[width=0.8in]{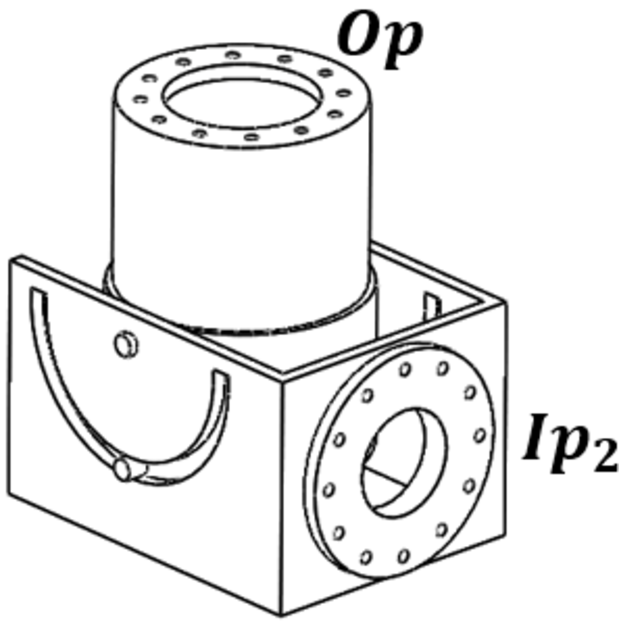}
	}
	~
	\subfigure[]{\includegraphics[width=1.1in]{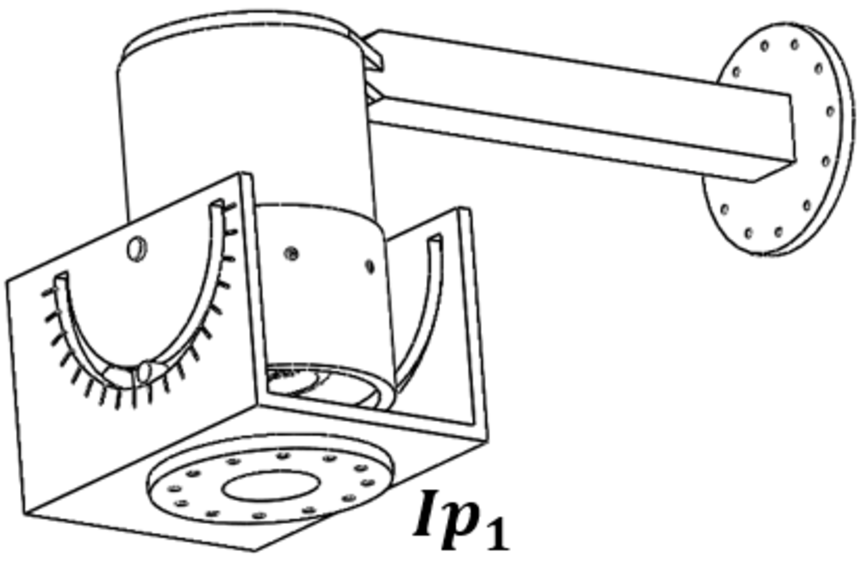}
	}
	~
	\subfigure[]{\includegraphics[width=1in]{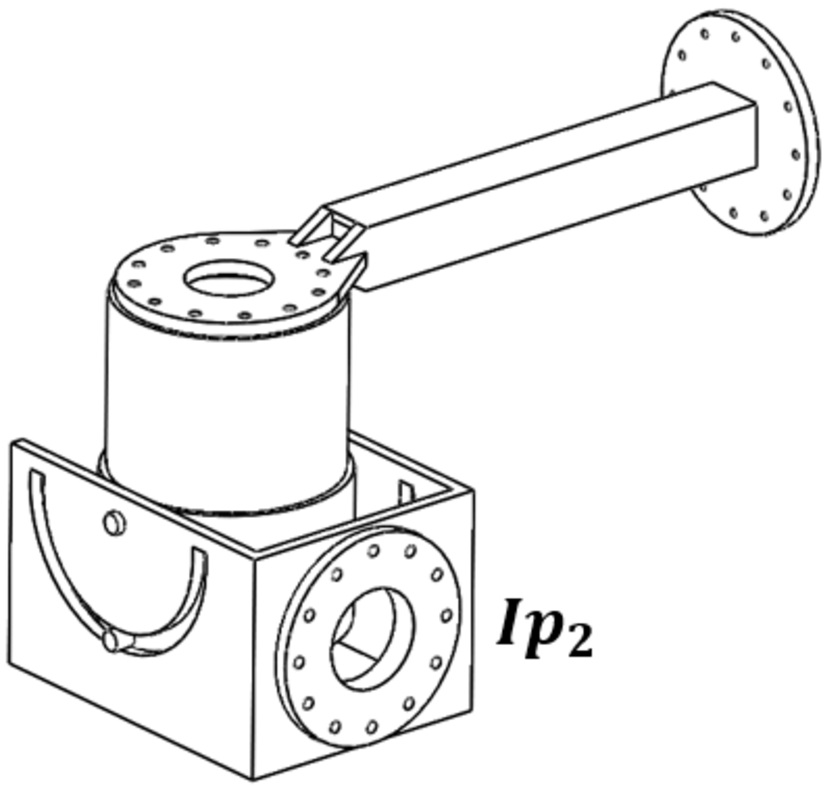}
	}
	\caption{The 4 types of modular units are as (a) $H^1$ or $L^1$, (b) $H^2$ or $L^2$, (c) $H^3$ or $L^3$ and (d) $H^4$ or $L^4$}
	\label{fig:4types}
\end{figure}

The proposed modular components can be assembled together to get the required custom configuration, even with unconventional twist angles. As there are two sizes of actuators used, two variants of modules are considered in the modular library as Heavy (H) and Light (L). For further assembly simplifications, four modular sub-assemblies (units) are proposed based upon the connection ports on the joint modules, as shown in Fig.~\ref{fig:4types}. These are valid for both `H and `L' variant of the joint modules. These can be enumerated as, `$H^k$' associated to Heavy (H) modules and `$L^k$' associated to Light (L) modules, where $k~\in~1:4$.
\begin{enumerate}
	\item `$H^1$' or `$L^1$' is used when a joint module is to be connected through $Ip_1$.
	
	\item `$H^2$' or `$L^2$' represents the corresponding joint module connected through $Ip_2$.
	
	\item `$H^3$' or `$L^3$' is used when a joint module is connected through $Ip_1$ and it is having a link module attached at $Op$. 
	
	\item `$H^4$' or `$L^4$' represents a joint  module connected through $Ip_2$ and having a link module at $Op$. 
\end{enumerate}
In this work, it has been assumed that the base modules or the first joint axis will always be `$H^1$'or `$H^3$' depending upon the spatial or planar case.
\subsection{Optimal Robotic Parameters to Modular Sequence and Combinations}\label{sec:Dh2modular}
The desired custom configuration can be realized by assembling the modules even without considering the DH parameters, using modular units sequence as desired. But the DH parameter approach provides a direct solution rather than trying out different combinations of modular units. DH parameters possess the information of number of degrees-of-freedom (DoF) and relationship between the adjacent joint frames in terms of the position and the orientation. As per the discussions in section~\ref{sec:modularconfigurations}, modular configurations are formed using four modular units. Therefore, first step of the process is to interpret a given set of DH parameters and convert them into modular units sequence i.e. in terms of $H^k ~or~ L^k$. Along with that, unconventional twist parameters are to be incorporated within these modular units, if any. The algorithm to convert DH parameters into modular unit sequence works upon the following points, these are valid for both H and L variants.
\begin{enumerate}
	\item If $a~\neq~0$ in the \textit{DH} table, use $H^3$ or $H^4$, else use $H^1$ or $H^2$.
	\item If $a=0~\cap~\alpha~\neq~0$ in the \textit{DH} table, twist is given by rotating actuator casings about the pivotal axis in the adaptive twist unit of joint module.
	\item If $a~\neq~0~\cap~\alpha~\neq~0$ in the \textit{DH} table, twist is given using $Ip_1$ or $Ip_2$ connections ports of the joint module and $H^3$ or $H^4$ are used.
	\item If $d~\neq~0$ in the \textit{DH} table, $H^1$ or $H^2$ are used.
\end{enumerate}
Through this, the prescribed DH parameters are converted into the modular unit sequence. To incorporate the different variants of the modular units (H or L), a strategy is proposed based upon the worst-torque analysis done by the authors in~\cite{dograJMD2021} to assemble the sequence.
\begin{enumerate}
	\item Fix link 1: $H^k$ and link \textit{n}: $L^k$.
	\item Connect $H^k$~$\rightarrow$ ~$H^k$ or $H^k$~$\rightarrow$ ~$L^k$.
	\item $L^k$~$\rightarrow$~ $L^k$.
	\item Refer $\epsilon$ for maximum number of each module carrying by the parent module in Table~\ref{tab:actuator_specs}
\end{enumerate}

Now considering this, for a given DoF and the value of $\epsilon$, there can be many possibilities of the combinations of the H and L modules. For e.g. 3-DoF configuration can be realized using $[L^k-L^k-L^k;~ H^k-L^k-L^k;~ H^k-H^k-L^k;~ H^k-H^k-H^k]$ combinations of modules. As maximum 4 number of H modules can be assembled in series, same is applicable to L, thus for a given DoF, the number of possible combinations can be deduced from
\begin{equation}
C_n = 5-|n-4|,~~~ n = DoF 
\end{equation} 
Therefore, for number of DoF = \{2, 3, 4, 5, 6, 7, 8\}, number of combinations comes out to be \{3, 4, 5, 4, 3, 2, 1\}. Selection of the modular assembly composition from the set of possible compositions becomes important so as to accomplish the task efficiently. This can be done using dynamic performance parameters such as minimized energy or torque required for the particular task. The selected modular composition is then modelled in a unified way for the motion planning and control of the realized modular manipulator through Robot Operating System (ROS). 

\section{Results and Discussions}\label{sec:results}
Constrained optimization as discussed in section~\ref{sec:opti}, is integrated with the modular units selection, formation of modular sequence.
\begin{figure}[htbp!]
	\centering
	
	\subfigure[]{\includegraphics[width=2in]{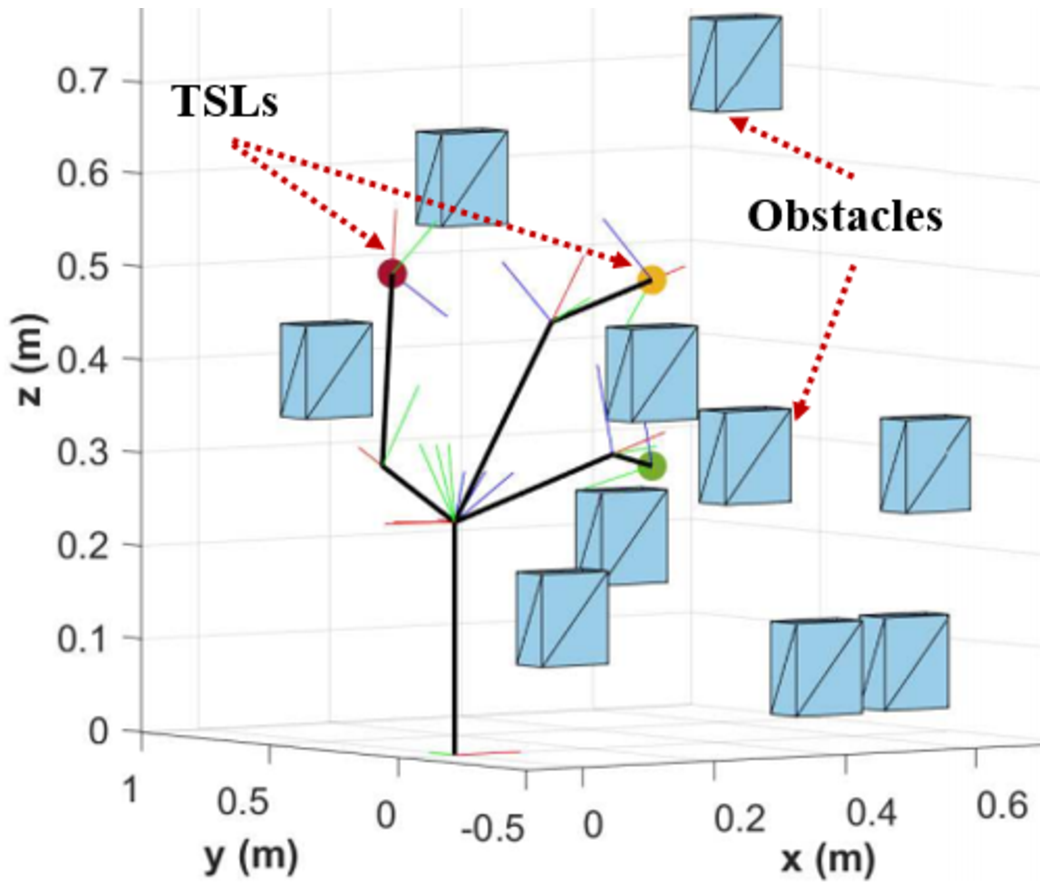}
	}
	~
	\subfigure[]{\includegraphics[width=1in]{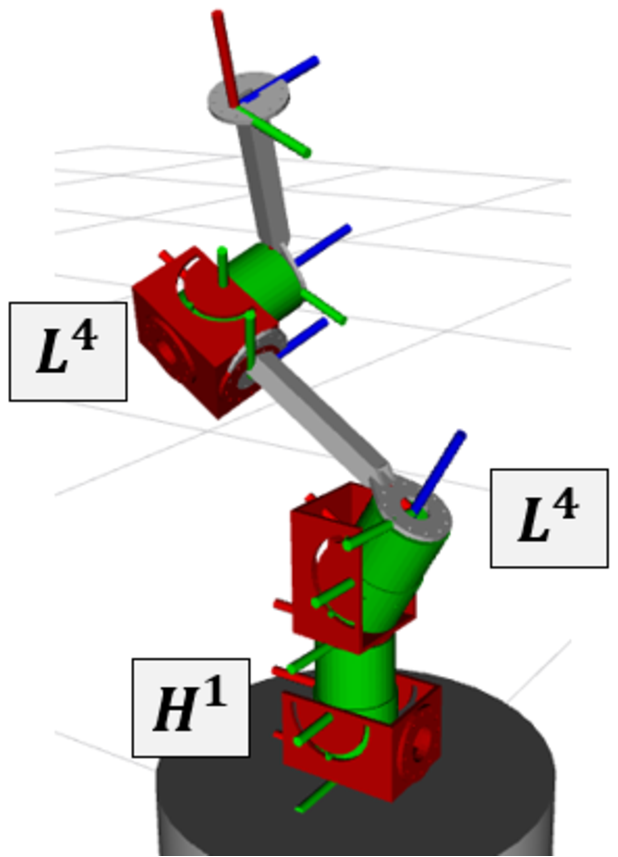}
	}
	~
	\subfigure[]{\includegraphics[width=1.4in]{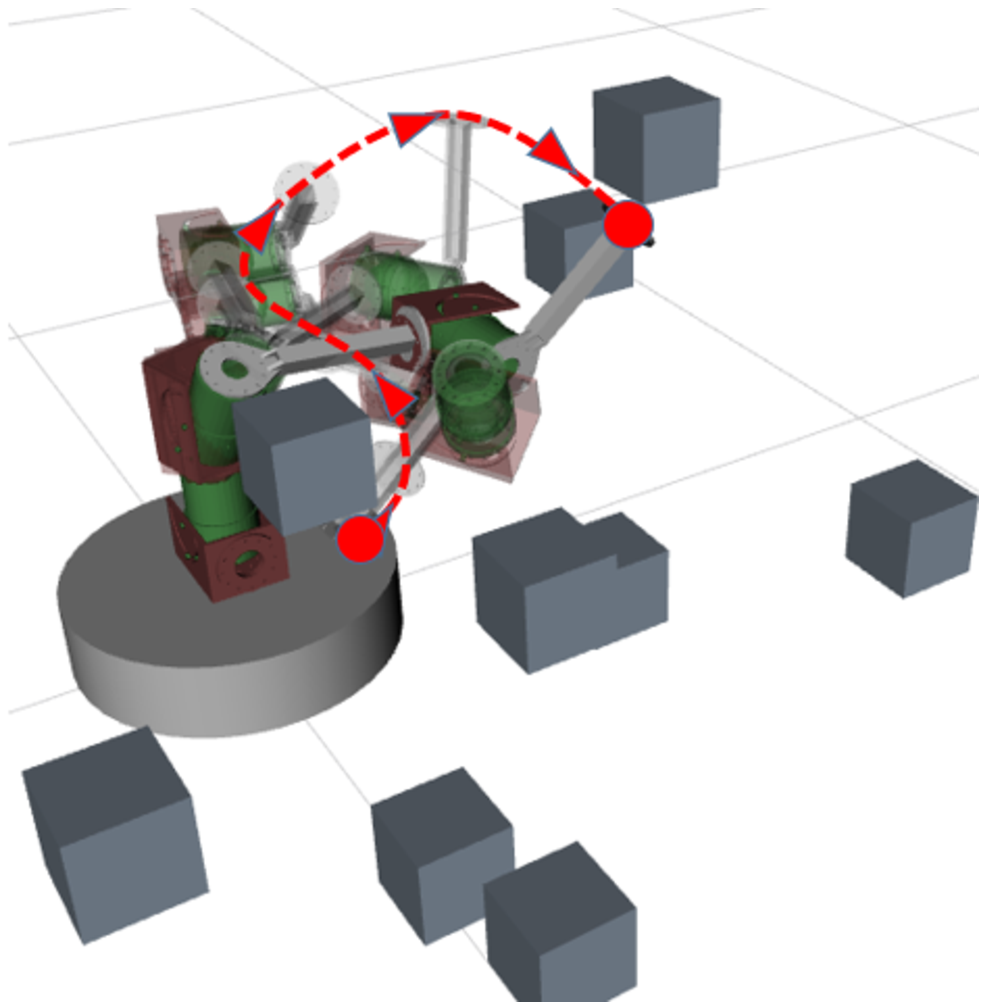}
	}
	~
	\caption{(a) 3-DoF optimal configuration reaching all 3-TSLs avoiding obstacles, (b) 3-DoF modular composition based upon synthesized configuration, (c) Motion planning between the TSLs of the modular composition using ROS.}
	\label{fig:test_result1}
\end{figure}

\begin{table}[h!]
	\caption{Optimal robotic parameters (DH) based upon the prescribed TSLs and cluttered environment.}
	\begin{center}
		\label{tab:DH_tsl}
		\setlength{\tabcolsep}{6pt}
		\begin{tabular}{c c c c c c c}
			& & & & & &\\
			\hline
			\hline\rule{0pt}{11pt}

			$i$ & $a_i$ (m) & $\alpha_i$ (rad)& $d_i$ (m) & $\theta_{i1}$ (rad) & $\theta_{i2}$ (rad) & $\theta_{i3}$ (rad)\\[2pt]

			\hline
			\hline\rule{0pt}{11pt}
			1 & 0 & 1.0367 & 0.25 & 2.6063 & 2.336 & 1.4483\\[2pt]
			2 & 0.3 & 1.0624 & 0 & 2.1362 & 2.8469 & 0.2030\\[2pt]
			3 & 0.3 & 0 & 0 & 1.7242 & 2.0784 & 0.6573\\ [2pt]

			\hline
			\hline
		\end{tabular}
	\end{center}
\end{table}
A test case is shown here to validate the algorithm discuss above. As shown in Fig.~\ref{fig:test_result1} (a), 10 obstacles are used as triangulated mesh cubes of sides $0.1~m$ each. These boxes are placed in a Cartesian space to make the environment cluttered. Three TSLs are chosen for reachability in between the boxes at (0.3, 0, 0.5), (0.3, 0, 0.3) and (0.1, 0.5, 0.5)\footnote{ All dimensions are in meters.}. The desired orientation parameters can be given in Euler angles format. Optimization problem in the inner loop, is solved by using Sequential Quadratic Program (SQP) with \textit{fmincon} function in MATLAB R2021a. Collision constraints are treated as non-linear inequality constraints for this problem, with $\delta=0.005$, and lower and upper limits on the design parameters are treated as bounds. 
Iterations are started with 5-DoF as per binary search algorithm, but eventually converged down to 3-DoF configuration to reach the prescribed working locations within the obstacles, as shown in Fig.~\ref{fig:test_result1} (a).  Table~\ref{tab:DH_tsl} shows the corresponding DH parameters of the optimal configuration and joint values for all the TSLs.

As briefed in section~\ref{sec:modularconfigurations}, the optimal parameters can be realized using modular architecture and modular units. The parameters are reduced to $H^1-L^4-L^4$ using given set of rules in section~\ref{sec:Dh2modular}, as shown in Fig.~\ref{fig:test_result1} (b). For the validation of the 3-DoF modular configuration, a Unified Robot Description Format (URDF) of the configuration is generated automatically to use it for the motion planning of the robot between the working locations avoiding obstacles. This has been implemented in Robot Operating System platform using Moveit~\cite{moveit}. Moveit uses a joint limited Jacobian based numerical solution for the inverse kinematics called as TracIK~\cite{tracik} and Open Motion Planning Library (OMPL)~\cite{ompl} for the motion planning of the end-effector of the manipulator under the constrained environment. The motion of the end-link of the 3-DoF configuration from working location 2 to 3 is implemented using this, as shown in Fig.~\ref{fig:test_result1} (c). This validates that the modular configuration is not only able to reach the working locations but also able to plan the motion between each task space location.   
\section{Conclusion}
Mapping of the optimal configuration synthesis of the manipulator to the modular composition with minimal number of DoF and considering unconventional parameters is presented in this paper. A nested bi-level constrained optimization problem is formulated with-respect-to prescribed working locations and given cluttered environment to synthesize a serial configuration with minimal number of DoF. A modular library is presented which can adapt to the unconventional robotic parameters to provide custom solutions based upon the user requirements. A strategy is presented to map the optimal unconventional configuration into the modular assembly with assembly possibilities and that is the novelty of the proposed integrated solution. A unified way of modeling and control of modular composition is presented using ROS, that validates the reachability and motion planning between the TSLs in a cluttered environment.

\bibliographystyle{spmpsci}
\bibliography{ms}

\begin{thebibliography}{10}
\providecommand{\url}[1]{{#1}}
\providecommand{\urlprefix}{URL }
\expandafter\ifx\csname urlstyle\endcsname\relax
  \providecommand{\doi}[1]{DOI~\discretionary{}{}{}#1}\else
  \providecommand{\doi}{DOI~\discretionary{}{}{}\begingroup
  \urlstyle{rm}\Url}\fi

\bibitem{tracik}
Beeson, P., Ames, B.: Trac-ik: An open-source library for improved solving of
  generic inverse kinematics.
\newblock In: 2015 IEEE-RAS 15th International Conference on Humanoid Robots
  (Humanoids), pp. 928--935. IEEE (2015).
\newblock \doi{10.1109/humanoids.2015.7363472}

\bibitem{Kinova2019}
Campeau-Lecours, A., Lamontagne, H., Latour, S., Fauteux, P., Maheu, V.,
  Boucher, F., Deguire, C., L'Ecuyer, L.J.C.: Kinova modular robot arms for
  service robotics applications.
\newblock In: Rapid Automation: Concepts, Methodologies, Tools, and
  Applications, pp. 693--719. IGI Global (2019).
\newblock \doi{10.4018/978-1-5225-8060-7.ch032}

\bibitem{campos2019task}
Campos, T., Inala, J.P., Solar-Lezama, A., Kress-Gazit, H.: Task-based design
  of ad-hoc modular manipulators.
\newblock In: 2019 International Conference on Robotics and Automation (ICRA),
  pp. 6058--6064. IEEE (2019).
\newblock \doi{10.1109/ICRA.2019.8794171}

\bibitem{chen2016modular}
Chen, I.M., Yim, M.: Modular robots.
\newblock In: Springer Handbook of Robotics, pp. 531--542. Springer (2016).
\newblock \doi{10.1007/978-3-319-32552-1_22}

\bibitem{moveit}
Coleman, D., Sucan, I., Chitta, S., Correll, N.: Reducing the barrier to entry
  of complex robotic software: a moveit! case study.
\newblock arXiv preprint arXiv:1404.3785  (2014)

\bibitem{dograRobotica2021}
Dogra, A., Padhee, S.S., Singla, E.: Optimal architecture planning of modules
  for reconfigurable manipulators.
\newblock Robotica pp. 1--15 (2021).
\newblock \doi{10.1017/S0263574720001174}

\bibitem{dograJMD2021}
Dogra, A., Sekhar~Padhee, S., Singla, E.: An optimal architectural design for
  unconventional modular reconfigurable manipulation system.
\newblock Journal of Mechanical Design \textbf{143}(6) (2021).
\newblock \doi{10.1115/1.4048821}

\bibitem{icer2017evolutionary}
Icer, E., Hassan, H.A., El-Ayat, K., Althoff, M.: Evolutionary cost-optimal
  composition synthesis of modular robots considering a given task.
\newblock In: Intelligent Robots and Systems (IROS), 2017 IEEE/RSJ
  International Conference on, pp. 3562--3568. IEEE (2017).
\newblock \doi{10.1109/iros.2017.8206201}

\bibitem{kereluk2017task}
Kereluk, J.A., Emami, M.R.: Task-based optimization of reconfigurable robot
  manipulators.
\newblock Advanced Robotics \textbf{31}(16), 836--850 (2017).
\newblock \doi{10.1080/01691864.2017.1362995}

\bibitem{liu2020optimizing}
Liu, S.B., Althoff, M.: Optimizing performance in automation through modular
  robots.
\newblock In: 2020 IEEE International Conference on Robotics and Automation
  (ICRA), pp. 4044--4050. IEEE (2020).
\newblock \doi{10.1109/icra40945.2020.9196590}

\bibitem{mohamed2017pose}
Mohamed, R.P., Xi, F.J., Chen, T.: A pose-based structural dynamic model
  updating method for serial modular robots.
\newblock Mechanical Systems and Signal Processing \textbf{85}, 530--555
  (2017).
\newblock \doi{10.1016/j.ymssp.2016.08.026}

\bibitem{singh2018modular}
Singh, S., Singla, A., Singla, E.: Modular manipulators for cluttered
  environments: A task-based configuration design approach.
\newblock Journal of Mechanisms and Robotics \textbf{10}(5), 051,010 (2018).
\newblock \doi{10.1115/1.4040633}

\bibitem{singh2016realization}
Singh, S., Singla, E.: Realization of task-based designs involving dh
  parameters: a modular approach.
\newblock Intelligent Service Robotics \textbf{9}(3), 289--296 (2016).
\newblock \doi{10.1007/s11370-015-0186-x}

\bibitem{stravopodis2020rectilinear}
Stravopodis, N., Moulianitis, V.: Rectilinear tasks optimization of a modular
  serial metamorphic manipulator.
\newblock Journal of Mechanisms and Robotics \textbf{13}(1) (2020).
\newblock \doi{10.1115/1.4047727}

\bibitem{ompl}
{\c{S}}ucan, I.A., Moll, M., Kavraki, L.E.: The {O}pen {M}otion {P}lanning
  {L}ibrary.
\newblock {IEEE} Robotics \& Automation Magazine \textbf{19}(4), 72--82 (2012).
\newblock \doi{10.1109/MRA.2012.2205651}

\bibitem{valente2016reconfigurable}
Valente, A.: Reconfigurable industrial robots: A stochastic programming
  approach for designing and assembling robotic arms.
\newblock Robotics and Computer-Integrated Manufacturing \textbf{41}, 115--126
  (2016).
\newblock \doi{10.1016/j.rcim.2016.03.002}

\bibitem{whitman2018task}
Whitman, J., Choset, H.: Task-specific manipulator design and trajectory
  synthesis.
\newblock IEEE Robotics and Automation Letters \textbf{4}(2), 301--308 (2018).
\newblock \doi{10.1109/lra.2018.2890206}

\bibitem{Modman2020}
Yun, A., Moon, D., Ha, J., Kang, S., Lee, W.: Modman: An advanced
  reconfigurable manipulator system with genderless connector and automatic
  kinematic modeling algorithm.
\newblock IEEE Robotics and Automation Letters \textbf{5}(3), 4225--4232
  (2020).
\newblock \doi{10.1109/lra.2020.2994486}

\end{thebibliography}

\end{document}